\documentclass[letterpaper, 10 pt, journal, twoside]{ieeeconf}

\IEEEoverridecommandlockouts

\usepackage{cite}
\usepackage{amsmath,amssymb,amsfonts}
\usepackage{graphicx}

\usepackage{float}
\usepackage{bm}
\usepackage{romannum}

\usepackage{booktabs}

\usepackage{algorithm}
\usepackage[noend]{algpseudocode}
\usepackage{textcomp}
\usepackage{dblfloatfix}

\usepackage[normalem]{ulem}
\usepackage[flushleft]{threeparttable}

\newcommand{\etal}{\textit{et~al}. }
\usepackage[font=small,labelfont=bf]{caption}

\usepackage{hyperref}
\usepackage[dvipsnames]{xcolor}
\newcommand\myshade{85}
\colorlet{mylinkcolor}{violet}
\colorlet{mycitecolor}{YellowOrange}
\colorlet{myurlcolor}{Aquamarine}
\hypersetup{
  linkcolor  = mylinkcolor!\myshade!black,
  citecolor  = mycitecolor!\myshade!black,
  urlcolor   = myurlcolor!\myshade!black,
  colorlinks = true,
}

\usepackage{balance}

\begin{document}
\title{\LARGE \bf Learning Multi-step Robotic Manipulation Policies from \\ Visual Observation of Scene and Q-value Predictions of Previous Action}


\author{Sulabh Kumra$^{1,2, *}$, Shirin Joshi$^{3}$ and Ferat Sahin$^{1}$
\thanks{$^{1}$ Rochester Institute of Technology, Rochester, NY, USA}%
\thanks{$^{2}$ OSARO Inc, San Francisco, CA, USA}%
\thanks{$^{3}$ Siemens Corporation, Corporate Technology, Berkeley, CA}
\thanks{{$^{*}$ Corresponding author. Email: \href{mailto:sk2881@rit.edu}{sk2881@rit.edu}}}
}





\markboth{IEEE Conference on Robotics and Automation (ICRA) 2022}
{Kumra \MakeLowercase{\textit{et al.}}: Learning Multistep Manipulation Policies} 

\maketitle

\begin{abstract}
In this work, we focus on multi-step manipulation tasks that involve long-horizon planning and considers progress reversal. Such tasks interlace high-level reasoning that consists of the expected states that can be attained to achieve an overall task and low-level reasoning that decides what actions will yield these states. We propose a sample efficient Previous Action Conditioned Robotic Manipulation Network (PAC-RoManNet) to learn the action-value functions and predict manipulation action candidates from visual observation of the scene and action-value predictions of the previous action. We define a Task Progress based Gaussian (TPG) reward function that computes the reward based on actions that lead to successful motion primitives and progress towards the overall task goal. To balance the ratio of exploration/exploitation, we introduce a Loss Adjusted Exploration (LAE) policy that determines actions from the action candidates according to the Boltzmann distribution of loss estimates. We demonstrate the effectiveness of our approach by training PAC-RoManNet to learn several challenging multi-step robotic manipulation tasks in both simulation and real-world. Experimental results show that our method outperforms the existing methods and achieves state-of-the-art performance in terms of success rate and action efficiency. The ablation studies show that TPG and LAE are especially beneficial for tasks like multiple block stacking. Additional experiments on Ravens-10 benchmark tasks suggest good generalizability of the proposed PAC-RoManNet.
\end{abstract}


\section{Introduction}
Robotic manipulation tasks have been the backbone of most industrial robotic applications, e.g. bin picking, assembly, palletizing, or machine tending operations. In structured scenarios these tasks have been reliably performed by the methods used in the existing work \cite{bohg2013data,kragic2003robust,kopicki2016one}. While, in unstructured scenarios, simple tasks such as pick only tasks have been successfully performed using grasping approaches such as \cite{levine2016end, redmon2015real, schmidt2018grasping, kumra2017robotic, yen2020learning, kumra2020antipodal}, complex tasks that involve multiple steps such as clearing a bin of mixed items, and creating a stack of multiple objects, still remain a challenge. While most prior manipulation methods focus on singular tasks, our work focuses on learning multi-step manipulation policies which generalize to a wide range of objects and tasks.

Deep reinforcement learning can be used to learn complex robotic manipulation tasks by using model-free deep policies. Kalashnikov \etal demonstrates this by proposing a QT-opt technique to provide a scalable approach for vision-based robotic manipulation applications \cite{kalashnikov2018scalable}. Riedmiller \etal introduced a SAC-X method that learns complex tasks from scratch with the help of multiple sparse rewards where only the end goal is specified \cite{riedmiller2018learning}. But training end-to-end manipulation policies that map directly from image pixels to joint velocities can be computationally expensive and time exhaustive due to a large volume of sample space and can be difficult to adapt on physical setups \cite{levine2018learning, joshi2020robotic, florence2019self, Rajeswaran-RSS-18}. To solve this, many have tried pixel-wise parameterization of both state and action spaces, which enables the use of a neural network as a Q-function approximator \cite{zeng2018learning, hundt2020good, berscheid2019robot}. However, these approaches have a low success rate, long learning time, and cannot handle intricate tasks consisting of multiple steps and long horizons.

\begin{figure}
    \centering
    \includegraphics[width=.98\linewidth]{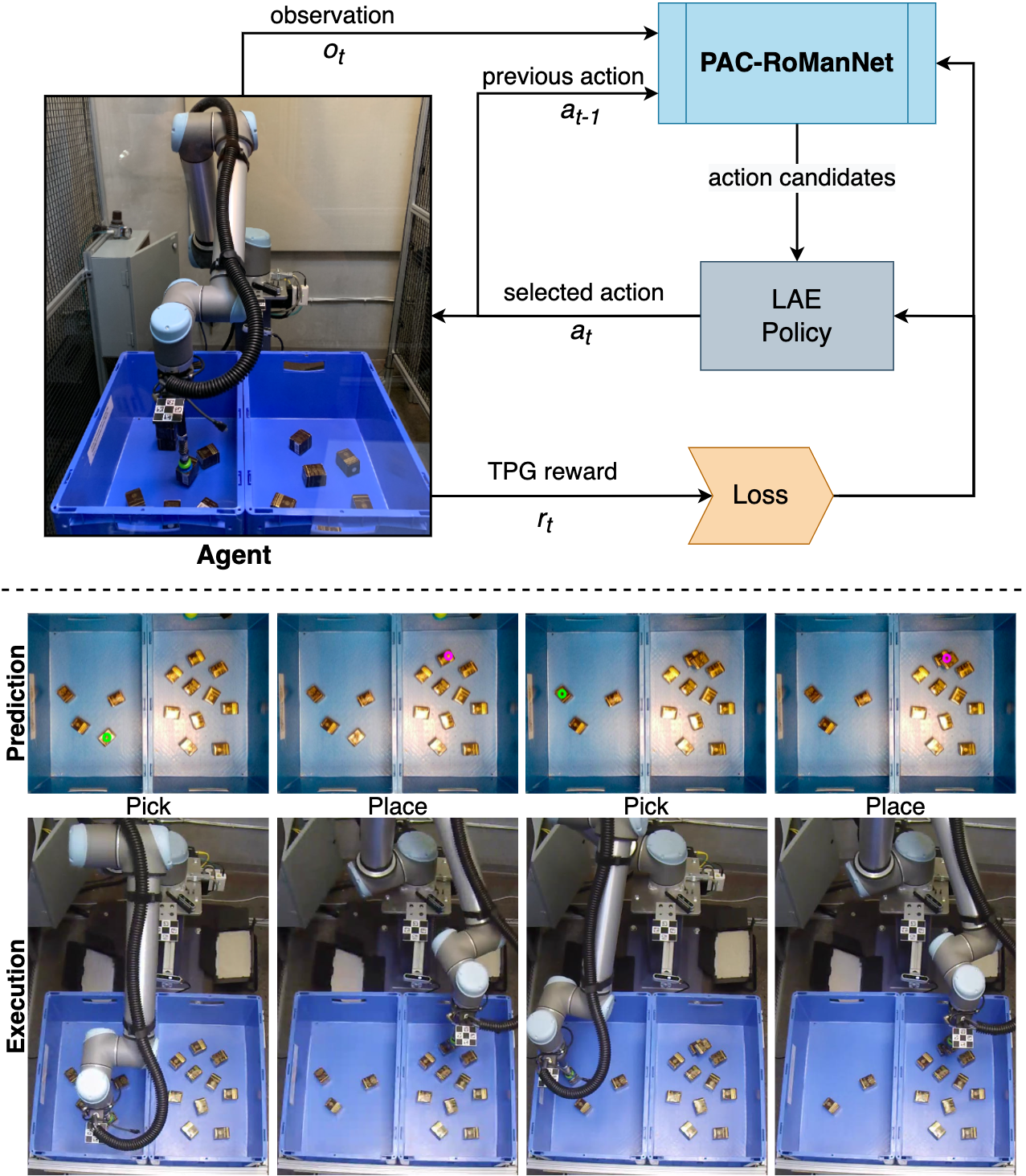}
    \caption{\textbf{Top}: Proposed approach for training an agent to efficiently learning multi-step manipulation tasks from visual observation of the scene and action-value predictions of the previous action. \textbf{Bottom}: Illustration of block stacking task being executed by the UR10 robot using the model trained for approximately 4 hours in the real world.}
    \label{fig: overview}
\end{figure}

Several prior works have proposed to use a model-free method to learn multi-step tasks through self-supervised learning. One such method proposed by Zeng \etal uses a Visual Pushing Grasping (VPG) framework that can discover and learn to push and grasp through model-free deep Q learning \cite{zeng2018learning}. Similarly, Jeong \etal performed a stacking task by placing a cube over another cube using a two-stage Self-Supervised Domain Adaptation (SSDA) technique \cite{jeong2020self}. Zhu \etal presented a framework in which manipulation tasks were learned by using a Deep Visuomotor Policy (DVP) that uses a combination of reinforcement learning and imitation learning to map RGB camera inputs directly into joint velocities \cite{zhu2018reinforcement}. Hundt \etal developed the Schedule for Positive Task (SPOT) framework \cite{hundt2020good}, that explores actions within the safety zones and can identify unsafe regions even without exploring and is able to prioritize its experience to only learn what is useful. Zeng \etal proposed Transporter Network trained using learning from demonstrations, which rearranges deep features to infer spatial displacements from visual input \cite{zeng2020transporter}. More closely related to our work is the VPG framework introduced by Zeng \etal which utilized a Fully Convolutional Network (FCN) as a function approximator to estimate the action-value function. However, this method had a low success rate and was sample inefficient.




We present a model-free deep reinforcement learning approach (shown in Fig. \ref{fig: overview}) to produce a deterministic policy that allows complex robot manipulation tasks to be effectively learned from pixel input. The policy directs a low-level controller to perform motion primitives rather than regressing motor torque vectors directly by learning a pixel-wise action success likelihood map. We propose a Task Progress based Gaussian (TPG) reward function to learn the coordinated behavior between intermediate actions and their consequences towards the advancement of an overall task goal. We introduce a Loss Adjusted Exploration (LAE) policy to explore the action space and exploit the knowledge, which helps in reducing the learning time and improving the action efficiency. The key contributions of this work are:
\begin{itemize}
    \item We introduce Previous Action Conditioned Robotic Manipulation Network (PAC-RoManNet), an end-to-end model architecture to efficiently learn the action-value functions and generate accurate action candidates from visual observation of the scene and Q-value predictions of the previous action.
    \item We propose a TPG reward function that uses a sub-task indicator function and an overall task progress function to compute the reward for each action in a multi-step manipulation task.
    \item We address the challenge of balancing the ratio of exploration/exploitation by introducing an LAE manipulation policy that selects actions according to the Boltzmann distribution of loss estimates.
\end{itemize}
We demonstrate the effectiveness of our approach in simulation as well as in the real-world setting by training an agent to learn three vision-based multi-step robotic manipulation tasks. PAC-RoManNet trained with TPG reward and LAE policy performed significantly better than previous methods with only 2000 iterations in the real-world setting. We observed a pick success rate of 93\% for a mixed item bin-picking task and 88\% action efficiency for a block-stacking task (illustrated in Fig. \ref{fig: overview}). We also demonstrate the generalizability of PAC-RoManNet by training it on the Ravens-10 benchmark tasks presented by Zeng \etal in \cite{zeng2020transporter} using a learning from demonstration framework.

\section{Problem Formulation}
\label{sec: probelm}

We consider the problem of efficiently learning multi-step robotic manipulation for unknown objects in an environment with unknown dynamics. Each manipulation task can be formulated as a Markov decision process where at any given state $s_t \in \mathcal{S}$ at time $t$, the robot makes an observation $o_t \in \mathcal{O}$ of the environment and executes an action $a_t \in \mathcal{A}$ based on policy $\pi(s_t)$ and receives an immediate reward of $\mathcal{R}_{a_t}( {s_t}, {s_{t+1}})$. In our formulation, $o_t \in \mathbb{R}^{4\times h \times w}$ is the visual observation of the robot workspace from RGB-D cameras, and we divide the action space $\mathcal{A}$ into two components: action type $\Phi$ and action pose $\Psi$. The underlying assumption is that the edges of $o_t$ are the boundaries of the agent’s workspace and $o_t$ embeds all necessary state information, thus providing sufficient information of the environment to choose correct actions.

\begin{figure*}
    \centering
    \vspace*{0.2cm}
    \includegraphics[width=0.98\textwidth]{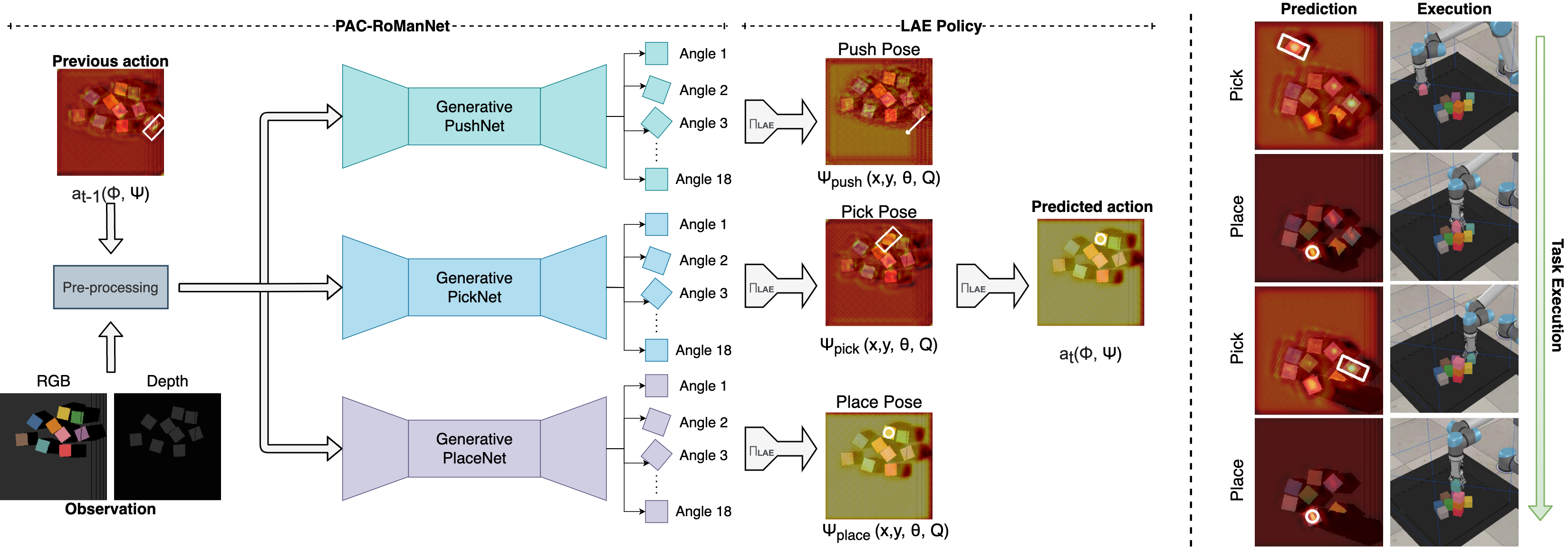}
    \caption{\textbf{Left}: Proposed PAC-RoManNet based framework for learning action-value function to predict manipulation action candidates from the observation of the state and Q value prediction of previous action. The LAE policy selects an action that maximizes the expected reward. \textbf{Right}: An illustration of an agent using the learned policy to execute a multi-step manipulation task, which requires the robot to create a stack of blocks with a goal stack height of 4. We can see the robot performing consecutive pick and place actions to build a stack using 10 cubes randomly placed in the bin. The robot learns not to pick blocks from the stack being built as it leads to progress reversal and thus a lower reward.}
    \label{fig: romannet}
\end{figure*}

The goal is to find an optimal policy $\pi^{*}$ in order to maximize the expected sum of future rewards i.e. $\gamma$-discounted sum on all future returns from time $t$ to $H$, across planning horizon $H$. An off-policy Q-learning can be used to train a greedy deterministic policy $\mathbf \pi({s_t})$ that chooses actions by maximizing the Q-function $ Q^\pi (s_t, a_t)$, which estimates the expected reward $\mathbb{E}_\pi(s_{t+1}, a_t)$ of taking action $a_t$ in state $s_t$ at time $t$.





\section{Proposed Approach}
\label{sec: approach}


We introduce PAC-RoManNet to approximate the action-value function $Q_\mu$, which predict manipulation action candidates $\mathcal{A}_{s_t}$ from the observation $o_t$ of the state $s_t$ at time $t$ and previous action $a_{t-1} (\Phi, \Psi)$. The proposed LAE manipulation policy $\Pi (s_t, \mathcal{L})$ determines the action $a_t (\Phi, \Psi)$ from the action candidates $\mathcal{A}_{s_t}$ that maximizes the reward $\mathcal{R}$. Once the agent executes $a_t (\Phi, \Psi)$, the reward is computed using a TPG reward function. The parameters $\mu$ are updated by minimizing the loss function. An overview of the proposed learning framework is illustrated in Fig. \ref{fig: romannet}.

\subsection{Manipulation Action Space}
We parameterize each manipulation action $a_t$ as two components: action type $\Phi$, which consists of three high-level motion primitives $\{push, pick, place\}$, and action pose $\Psi$, which is defined by the pose at which the action is performed. Each manipulation action pose in image frame $\Psi$ is parameterized pixel-wise and defined as:
\begin{equation}
    \Psi = (x, y, \theta, Q)
\end{equation}
where $(x, y)$ is the center of action pose in image coordinates, $\theta$ is the rotation in the image frame, and $Q$ is an affordance score that depicts the  “quality” of action.    

The high-level motion primitive behaviors $\Phi$ are defined as follows:
\begin{itemize}
    \item \textbf{Pushing:} $\Psi_{push} = (x, y, \theta, Q)$ denotes the starting pose of a 10cm push. A push starts with the gripper closed at $(x, y)$ and moves horizontally at a fixed distance along angle $\theta$.
    \item \textbf{Picking:} $\Psi_{{pick}} = (x, y, \theta, Q)$ denotes the middle position of a top-down grasp. During a pick attempt, both fingers attempt to move 3cm below $\Psi_{{pick}}$ (in the $-z$ direction) before closing the fingers.
    \item \textbf{Placing:} $\Psi_{place} = (x, y, \theta, Q)$ denotes the middle position of a top-down placement. During a place attempt, both fingers open when the place pose $\Psi_{place}$ is reached.
\end{itemize}

\subsection{Learning the Action-Value Functions}
We approximate the action-value function using a neural network architecture we call PAC-RoManNet. It consists of three Generative Residual Convolutional Neural Network (GR-ConvNet) \cite{kumra2020antipodal} models PushNet, PickNet, and PlaceNet, one for each motion primitive behavior (pushing, picking, and placing respectively). Each individual GR-ConvNet model takes as input the image representation $o_t$ of the state $s_t$ and Q value prediction of previous action $a_{t-1} (\Phi, \Psi)$, and generates a pixel-wise map $Q(s_t, a_t) \in \mathbb{R}^{h \times w}$, where Q value prediction at each pixel represents the future expected reward $\mathbb{E}_\pi(s_{t+1}, a_t)$ of executing action $a_t (\Phi, \Psi)$. The action that maximizes the Q-function is the action type $\Phi$ and the pixel with the highest Q value is the pose $\Psi$.

Instead of regressing the gripper position and orientation as in \cite{kumra2021learning}, in this work, we discretize the orientations into 18 angles (0\textdegree, 20\textdegree, 40\textdegree, .., 340\textdegree) by rotating the input image representation $o_t$ of the state $s_t$. The output of each motion primitive network is a set of 18 images $\mathbb{R}^{h \times w \times 18}$, which depict the action success scores for each of the 18 angles.

The PAC-RoManNet is continuously trained to approximate the optimal policy with prioritized experience replay using stochastic rank-based prioritization and leverages future knowledge via a recursively defined expected reward function:
\begin{equation}
    \mathbb{E}_\pi(s_{t+1}, a_t) = 
    \mathbb{R}(s_{t+1}, a_t) + \eta (\gamma \mathbb{R}(s_{t+2}, a_{t+1}))
\end{equation}
\noindent where $\eta$ is a reward propagation factor, which equals to $1$ if $\mathbb{R}(s_{t+1}, a_t) > 0$ and $0$ otherwise. This ensures that future rewards only propagate across time steps where subtasks are completed successfully.


\subsection{Task Progress based Gaussian Reward}
The reward function $\mathcal{R}(s_{t+1}, a_t) \in \mathbb{R}^{h \times w}$ operates on two principles: actions which advance overall task progress receive a reward proportional to the quantity of progress, but actions which reverse the progress receive 0 reward. The task progress is measured using: (\romannum{1}) a sub-task indicator function $\mathcal{X}(s_{t+1}, a_t)$, which equals to $1$ if $a_t$ leads to a successful primitive action and $0$ otherwise, and (\romannum{2}) an overall task progress function $\mathcal{P}(s_{t+1}, a_t) \in [0, 1]$, which is proportional to the progress towards an overall goal. We define our task progress based reward function as:
\begin{equation}
    \mathcal{R}_{tp}(s_{t+1}, a_t) = \mathcal{W}(\Phi) \mathcal{X}(s_{t+1}, a_t) \mathcal{P}(s_{t+1}, a_t)
\end{equation}
\noindent where $\mathcal{W}(\Phi)$ is a weighting function that depends on the primitive motion action type $\Phi$. In this work, a push action is successful if it perturbs an object, a pick action is successful if an object is grasped and raised from the surface, and a place action is successful only if it increases the stack height.

The reward is a scalar spike and we smooth it using an anisotropic Gaussian distribution \cite{tsiotsios2013choice} parameterized with standard deviations $\sigma_x$ and $\sigma_y$. We used an anisotropy ratio of 2 (i.e., $\sigma_x / \sigma_y = 2$), where x is aligned with the $x$-axis of the gripper. This new Task Progress based Gaussian (TPG) reward is specified as follows:
\begin{equation}
    G(x, y, \sigma_x, \sigma_y) = 
    \frac{1}{{2\pi \sigma_x \sigma_y}}e^{{ - \left( \frac{x^2}{2 \sigma_x ^2} + \frac{y^2}{2 \sigma_y ^2} \right) }}
\end{equation}
\begin{equation}
    \mathcal{R}_{g}(s_{t+1}, a_t) = \mathcal{R}_{tp}(s_{t+1}, a_t) \circledast G(x, y, \sigma_x, \sigma_y)
\end{equation}
\begin{equation}\label{eq: reward}
    \mathcal{R}_{tpg}(s_{t+1}, a_t) = \max ( \mathcal{R}_{tp}(s_{t+1}, a_t), \mathcal{R}_{g}(s_{t+1}, a_t))
\end{equation}
where $\circledast$ is the convolution operator, and $G$ is the anisotropic Gaussian filter applied to $\mathcal{R}_{tp}$.

\subsection{Loss Adjusted Exploration Policy}
To reduce unnecessary exploration once knowledge about initial states has been sufficiently established, we introduce Loss Adjusted Exploration (LAE) to extend the $\epsilon$-greedy policy. We define the update steps for loss dependent exploration probability $\mathcal{E} (\mathcal{L})$ as the following Boltzmann distribution of loss estimates:
\begin{equation}
    f(\mathcal{L}_t, \sigma) = \frac{1-e^(\frac{-|\alpha \cdot \mathcal{L}_t|}{\sigma})}
    {1+e^(\frac{-|\alpha \cdot \mathcal{L}_t|}{\sigma})}
\end{equation}

\begin{equation}
    \mathcal{E}_{t+1} (\mathcal{L}_t) = \beta \cdot f(\mathcal{L}_t, \sigma) + (1-\beta) \cdot \mathcal{E}_t (\mathcal{L}_t)
\end{equation}
 
\noindent where $\mathcal{L}_t$ is the loss of the network computed using a robust loss function \cite{barron2019general} at each time step $t$, $\sigma$ is a positive constant called inverse sensitivity and $\beta \in [0,1)$ is a parameter determining the influence of the selected action on the exploration rate. The LAE policy is defined as:
\begin{equation}\label{eq: policy}
    \Pi_{LAE}({s_t}, \mathcal{L}_t) = \begin{cases}
    \mathcal{U} (\mathcal{A}_{s_t}) & \text{if } \xi < \mathcal{E} (\mathcal{L}), \\
    \operatorname*{arg\,max} _ { a_t \in \mathcal{A}(s_t)} Q_\mu (s_t, a_t)              & \text{otherwise.}
    \end{cases}
\end{equation}
\noindent where $\mathcal{U}$ is a uniform distribution over action candidates $\mathcal{A}_{s_t}$ and $\xi \in [0,1)$ is a random number drawn at each time step $t$ from a uniform distribution.


\section{Experiments}

We conduct experiments in both simulated and real settings to evaluate the proposed method across various tasks. We design our experiments to investigate the following three questions: (\romannum{1}) How well does our method perform on different multi-step manipulation tasks? (\romannum{2}) Does our method improve task performance as compared to baseline methods? (\romannum{3}) What are the effects of the individual components of our framework in solving multi-step manipulation tasks, i.e., without the Loss Adjusted Exploration or Task Progress-based Gaussian reward?


\subsection{Experimental Setups}
For the simulated environment, we expanded on the CoppeliaSim based setup used in \cite{zeng2018learning} to provide a consistent environment for fair comparisons and ablations. The environment simulates the agent using a UR5 robot arm with an RG2 gripper. Bullet Physics 2.83 is used to simulate the dynamics and CoppeliaSim’s internal inverse kinematics module for robot motion planning. A statically mounted perspective 3D camera is simulated in the environment to capture the observations of the states. For the real-world setup, we used a UR10 robot with a Piab suction cup as the end effector. The RGB-D images of the scene are captured using an Intel RealSense D415 mounted rigidly above the workspace.

\subsection{Evaluation Metrics}
For evaluating the trained model, the policy is greedy deterministic and the model weights are reset to the trained weights at the start of each new test run. For each of the test cases, we execute 30 runs with new random seeds and evaluate performance with the following metrics found in \cite{zeng2018learning, hundt2020good}:
\begin{itemize}
    \item Completion rate: the average percentage runs in which the policy completed the given task without 10 consecutive fail attempts.
    \item Pick success rate: the average percentage of object picking success per completion.
    \item Action efficiency: a ratio of the ideal to the actual number of actions taken to complete the given task.
\end{itemize}



\subsection{Simulation Experiments}
We design three manipulation tasks to evaluate our method in simulation: dense clutter removal, clearing 11 challenging test cases with adversarial objects, and stacking multiple objects. All tasks share the same MDP formulation in section \ref{sec: probelm}, while the object set and the reward function are different for each task. For fair comparison with prior work, the same sets of objects as in VPG \cite{zeng2018learning} and SPOT \cite{hundt2020good} are used in these tasks. To train our models for all tasks, objects with randomly selected shapes and colors are dropped in front of the robot at the start of each experiment. The robot then automatically performs data collection by trial and error. For object removal-related tasks in simulation, objects are removed from the scene after a successful pick action. The environment is reset at the termination of the task.

\begin{table}
\begin{center}
\caption{Performance for Dense Clutter Removal Task (Mean \%)}
\label{tab:Random_Arrangements}
\begin{tabular}{l c c c}
\toprule
\textbf{Approach} & \textbf{Completion} & \textbf{Pick} & \textbf{Action}\\
 & \textbf{Rate} & \textbf{Success} & \textbf{Efficiency}\\
\midrule
Grasping-only \cite{zeng2018robotic} & 90.9 & 55.8 & 55.8 \\
P+G Reactive \cite{zeng2018learning} & 54.5 & 59.4 & 47.7 \\
VPG \cite{zeng2018learning} & 100 & 67.7 & 60.9 \\
SPOT \cite{hundt2020good} & 100 & 84 & 74 \\
\midrule
Ours & \textbf{100} & \textbf{96.2} & \textbf{94.6} \\
\bottomrule
\end{tabular}
\end{center}
\end{table}

\textbf{Dense Clutter Removal Task:}
We first evaluate the proposed method in the simulated environment where 30 objects are randomly dropped in a bin. The goal of the agent is to remove all objects from the bin in front of the robot. We trained the agent with only 10 objects instead of 30 objects. This helps to test the generalization of policies to more cluttered scenarios. The first experiment compares the proposed method to previous methods in simulation using the adversarial objects from \cite{zeng2018learning}. Comparison of our results with Grasping-only \cite{zeng2018robotic}, P+G Reactive \cite{zeng2018learning}, VPG \cite{zeng2018learning} and SPOT \cite{hundt2020good} are summarized in Table \ref{tab:Random_Arrangements}. We see that with a pick success rate of 96.2\%, the proposed method outperforms all other methods across all metrics. We observe that our proposed method learned to pick items in a dense clutter rather than trying to declutter them by pushing before picking. We hypothesize that this led to the high action efficiency of 94.6\%.

\begin{table}
\begin{center}
\caption{Performance for Challenging Arrangements (Mean \%)}
\label{tab:Challenging_Arrangements}
\begin{tabular}{lccc}
\toprule
\textbf{Approach} & \textbf{Cases 100\%} & \textbf{Completion} & \textbf{Action}\\
 & \textbf{Completed} & \textbf{Rate} & \textbf{Efficiency}\\
\midrule
Grasp-only \cite{zeng2018robotic} & - & 40.6 & 51.7 \\
P+G Reactive \cite{zeng2018learning} & - & 48.2 & 46.4 \\
VPG \cite{zeng2018learning} & 5/11 & 82.7 & 60.1 \\
SPOT \cite{hundt2020good} & 7/11 & 95 & 38 \\
\midrule
Ours & \textbf{10/11} & \textbf{98.8} & \textbf{89.4} \\
\bottomrule
\end{tabular}
\end{center}
\end{table}

\textbf{Challenging Arrangements Task:}
We also compare the proposed method with other methods in simulation on 11 challenging test cases with adversarial clutter from \cite{zeng2018learning}. These test cases are manually engineered to reflect challenging picking scenarios which allow us to evaluate the model’s robustness. Each test case consists of a unique configuration of 3 to 6 objects placed in a tightly packed arrangement that will be challenging even for an optimal picking-only policy as it will require de-cluttering them prior to picking. As a sanity check, a single isolated object is additionally placed separately from the main configuration. Similar to the dense clutter removal task, we trained a policy that learns to push and pick with 10 objects randomly placed in a bin and tested it on 11 challenging test cases. Performance comparison with previous work is shown in Table \ref{tab:Challenging_Arrangements}. Across the collection of test cases, we observe that our proposed method can successfully solve 10/11 cases with a 100\% completion rate and an overall completion rate of 98.6\%. Moreover, the action efficiency of 89.4\% with our method indicates that our method is significantly better than VPG \cite{zeng2018learning} and SPOT \cite{hundt2020good} methods that attained action efficiencies of only 60.1\% and 38\% respectively.

\begin{table}
\begin{center}
\caption{Comparison of performance for block stacking task in terms of mean \% action efficiency for various goal stack heights}
\label{tab:block_stacking}
\begin{tabular}{lcccc}
\toprule
\textbf{Approach} & \textbf{Stack of 2} & \textbf{Stack of 3} & \textbf{Stack of 4} & \textbf{Stack of 5} \\
\midrule
DVP \cite{jeong2020self} & 35 & - & - & - \\
SSDA \cite{zeng2018learning} & 62 & - & - & - \\
VPG \cite{zeng2018learning} & 53 & 32 & 12 & 4 \\
SPOT \cite{hundt2020good} & 74 & 61 & 48 & 35 \\
\midrule
Ours & \textbf{99} & \textbf{96} & \textbf{94} & \textbf{91} \\
\bottomrule
\end{tabular}
\end{center}
\end{table}

\textbf{Block Stacking Task:}
To truly evaluate our multi-step task learning method, we consider the task of stacking multiple blocks on top of each other. This constitutes a challenging multi-step robotic task as it requires the agent to acquire several core abilities: picking a block from 10 blocks arbitrarily placed in the bin, precisely placing it on top of the second block, and repeating this process until a goal stack height is reached. We performed multiple experiments with a goal stack height in the range of 2 to 5. Table \ref{tab:block_stacking} summarises the performance of our method compared to prior work. Jeong \etal used self-supervised domain adaptation (SSDA) \cite{jeong2020self} and Zhu \etal used deep visuomotor policy (DVP) \cite{zhu2018reinforcement} and tested it with a goal stack height of 2. Hundt \etal used SPOT \cite{hundt2020good} and tested it with a goal stack height of 4. For this task with the highest complexity, our approach seems to perform significantly better with the action efficiency of 99\% and 94\% for a goal stack height of 2 and 4 respectively. A possible reason is that, although the task is very complex, the proposed PAC-RoManNet and TPG reward function helps learn optimal multi-step manipulation policy.

\begin{table}
\begin{center}
\caption{Comparison of performance of state-of-the-art reinforcement learning based grasping methods in real-world settings}
\label{tab: real-grasping}
\begin{tabular}{l c c c}
\toprule
\textbf{Approach} & \textbf{Success \%} & \textbf{Training Steps} & \textbf{Test Items}\\
\midrule
VPG \cite{zeng2018learning} & 68 & 2.5k & 20 seen\\
SPOT \cite{hundt2020good} & 75 & 1k & 20 seen \\
Levine \etal \cite{levine2018learning} & 78 & 900k & - \\
QT-Opt \cite{kalashnikov2018scalable} & 88 & 580k & 28 seen \\
Grasp-Q-Network \cite{joshi2020robotic} & 89 & 7k & 9 seen \\
Berscheid \etal \cite{berscheid2019robot} & 92 & 27.5k & 20 seen\\
\midrule
Ours & \textbf{96 $\pm$ 1} & 2k & 20 seen \\
 & \textbf{93 $\pm$ 2} & 2k & 10 unseen \\
\bottomrule
\end{tabular}
\end{center}
\end{table}



\begin{table*}
\begin{center}
\vspace*{0.2cm}
\setlength\tabcolsep{4pt}
\caption{PAC-RoManNet performance on Ravens-10 benchmark tasks. Task success rate (mean \%) vs demonstration used in training.}
\label{tab: Benchmark}
\begin{tabular}{l cccc cccc cccc cccc cccc}
\toprule
 & \multicolumn{4}{c}{align-box-corner} &  \multicolumn{4}{c}{assembling-kits} & \multicolumn{4}{c}{block-insertion} & \multicolumn{4}{c}{manipulating-rope} & \multicolumn{4}{c}{packing-boxes} \\

\cmidrule(lr){2-5} \cmidrule(lr){6-9} \cmidrule(lr){10-13} \cmidrule(lr){14-17} \cmidrule(lr){18-21}

Approach & 1 & 10 & 100 & 1000 & 1 & 10 & 100 & 1000 & 1 & 10 & 100 & 1000 & 1 & 10 & 100 & 1000 & 1 & 10 & 100 & 1000 \\

\midrule

PAC-RoManNet &
\textbf{62.0} & \textbf{91.0} & \textbf{100} & \textbf{100} &
\textbf{56.8} & \textbf{83.2} & \textbf{99.4} & \textbf{99.6} &
\textbf{100} & \textbf{100} & \textbf{100} & \textbf{100} &
\textbf{25.7} & \textbf{87.0} & \textbf{99.9} & \textbf{100} &
\textbf{96.1} & \textbf{99.9} & \textbf{99.9} & \textbf{100} \\

Transporter\cite{zeng2020transporter}&
35.0 & 85.0 & 97.0 & 98.3 &
28.4 & 78.6 & 90.4 & 94.6 &
\textbf{100} & \textbf{100} & \textbf{100} & \textbf{100} & 
21.9 & 73.2 & 85.4 & 92.1 & 
56.8 & 58.3 & 72.1 & 81.3 \\

Form2Fit\cite{zakka2020form2fit} &
7.0 & 2.0 & 5.0 & 16.0 & 
3.4 & 7.6 & 24.2 & 37.6 &
17.0 & 19.0 & 23.0 & 29.0 &
11.9 & 38.8 & 36.7 & 47.7 &
29.9 & 52.5 & 62.3 & 66.8 \\

\midrule

 & \multicolumn{4}{c}{palletizing-boxes} &  \multicolumn{4}{c}{place-red-in-green} & \multicolumn{4}{c}{stack-block-pyramid} & \multicolumn{4}{c}{sweeping-piles} & \multicolumn{4}{c}{towers-of-hanoi} \\

\cmidrule(lr){2-5} \cmidrule(lr){6-9} \cmidrule(lr){10-13} \cmidrule(lr){14-17} \cmidrule(lr){18-21}

 & 1 & 10 & 100 & 1000 & 1 & 10 & 100 & 1000 & 1 & 10 & 100 & 1000 & 1 & 10 & 100 & 1000 & 1 & 10 & 100 & 1000 \\

\midrule

PAC-RoManNet &
\textbf{84.2} & \textbf{98.2} & \textbf{100} & \textbf{100} & 
\textbf{92.3} & \textbf{100} & \textbf{100} & \textbf{100} & 
\textbf{23.5} & \textbf{79.3} & \textbf{94.6} & \textbf{97.1} & 
\textbf{98.2} & \textbf{99.1} & \textbf{99.2} & \textbf{98.9} &
\textbf{98.2} & \textbf{99.9} & \textbf{99.9} & \textbf{99.9} \\

Transporter\cite{zeng2020transporter}&
63.2 & 77.4 & 91.7 & 97.9 &
84.5 & \textbf{100} & \textbf{100} & \textbf{100} &
13.3 & 42.6 & 56.2 & 78.2 &
52.4 & 74.4 & 71.5 & 96.1 &
73.1 & 83.9 & 97.3 & 98.1 \\

Form2Fit\cite{zakka2020form2fit} &
21.6 & 42.0 & 52.1 & 65.3 &
83.4 & \textbf{100} & \textbf{100} & \textbf{100} &
19.7 & 17.5 & 18.5 & 32.5 &
13.2 & 15.6 & 26.7 & 38.4 &
3.6 & 4.4 & 3.7 & 7.0 \\

\bottomrule
\end{tabular}
\end{center}
\end{table*}

\subsection{Real-World Experiments}
We validate our method in the real world on two tasks: mixed item bin picking and block stacking. Fig. \ref{fig: overview} illustrates the block stacking task being executed by the UR10 robot using PAC-RoManNet trained for 2k iterations ($\sim4$ hours of robot run time). We provide additional qualitative results in the attached video.

\textbf{Mixed Item Bin Picking:} For the mixed item bin-picking task, the robot picks items from the source bin and places them into a destination bin until the source bin is empty, and then swaps the source and destination bin for the next run. 30 different items were used for training the model. To automate the training process, suction feedback is used to detect a successful pick and a binary classifier is used to detect if the bin is empty after each manipulation action. We observe that the robot adopts diverse strategies to clear the clutter. The performance results in Table \ref{tab: real-grasping} shows that PAC-RoManNet performs consistently better than all previous methods. Pick success rate of 96\% after training for only 2k iterations demonstrates the high sample efficiency of our method. Generalization is a key index for applications in industrial and logistics automation. Our experiments with 10 unseen items show that our method is extremely impressive as it can generalize to novel objects in a real-world setting and still achieve a success rate of 93\%.


\textbf{Real-World Block Stacking:} The real-world block stacking task is similar to the one in simulation, where the robot performs manipulation actions until the goal stack height is reached. The depth measurements from the overhead camera are used to determine the current stack height. This task is particularly challenging as the agent needs to learn to align the position and orientation of the picked item with the stack during placement. Remarkably, we observed a 100\% completion rate and 88\% action efficiency for a goal stack height of 4, outperforming the state-of-the-art action efficiency of 61\% reported by Hundt \etal in \cite{hundt2020good}.

\subsection{Ablation Studies}
\begin{figure}
    \centering
    \includegraphics[width=1\linewidth]{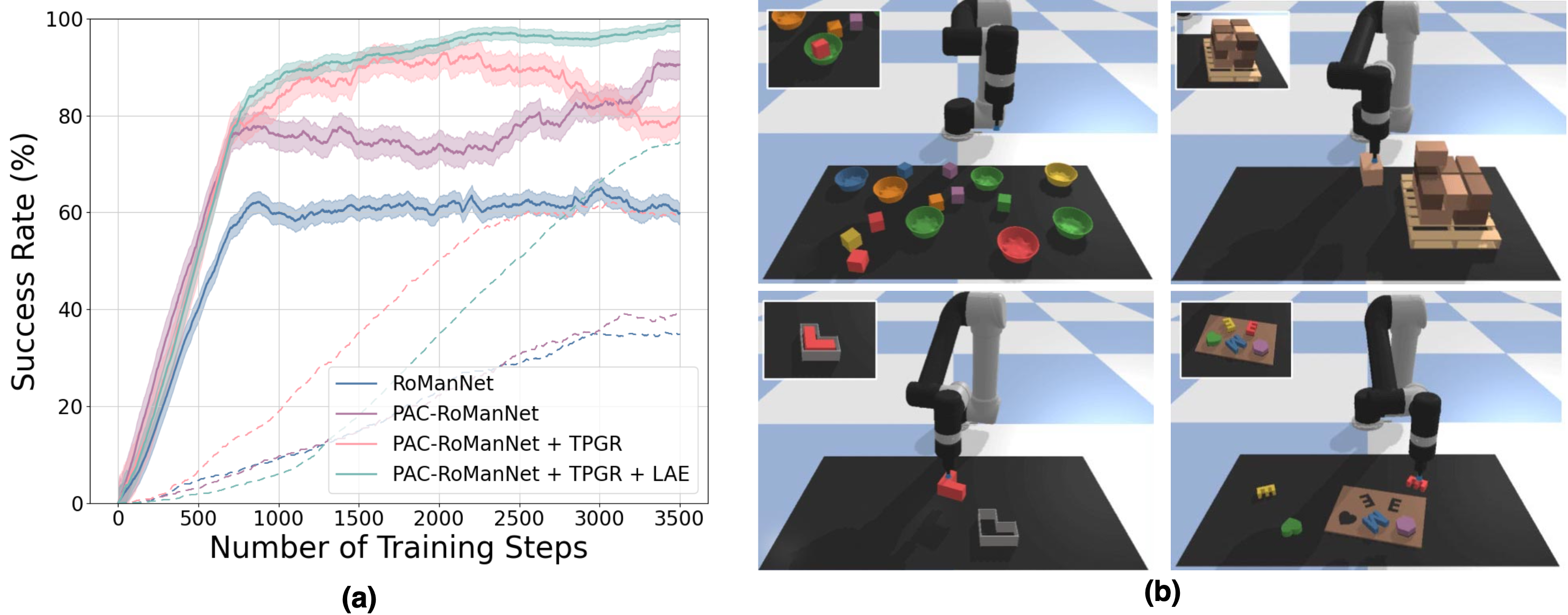}
    \caption{(a) Learning curves for ablation of techniques used in conjunction with PAC-RoManNet for training agent on block stacking task. Solid lines indicate mean action success rates and dotted lines indicate mean action efficiency over training steps. (b) Examples of Ravens-10 benchmark tasks. Left to right row-major order: place-red-in-green, palletizing-boxes, block-insertion, assembling-kits.}
    \label{fig: ablation}
\end{figure}

In order to assess the necessity and efficacy of the different components of our proposed method, described in section \ref{sec: approach}, we provide ablation experimental results. Fig. \ref{fig: ablation}(a) compares the learning curves for the underlying algorithm against the baseline approaches for the complex multi-step task of block stacking.

\textbf{PAC-RoManNet}: We establish a baseline using the primary reward function found in VPG \cite{zeng2018learning} and decaying $\epsilon$-greedy exploration strategy. We define the baseline reward function as a function of sub-task indicator only, i.e $\mathcal{R}_{base}(s_{t+1}, a_t) = \mathcal{X}(s_{t+1}, a_t)$. The only case where this baseline model is on par with the full model is the clutter removal task, in which both the baseline and the full model achieved similar levels of performance. We hypothesize that this is due to the short length of the task, where the task progress-based reward did not play a significant role.

\textbf{PAC-RoManNet + TPGR}: The second baseline is a combination of the TPG reward function and $\epsilon$-greedy exploration strategy. We observe that TPG reward helped the agent learn to avoid task progress reversal for the multi-step block stacking task, thus improving the success rate by more than 25\%. We also observed a significant improvement in action efficiency for the challenging arrangements and block stacking tasks.

\textbf{PAC-RoManNet + TPGR + LAE}: For the complete proposed model, the LAE strategy is used instead of the $\epsilon$-greedy exploration. We observe that the LAE policy helps in further increasing the action efficiency by 11\% for the block stacking task as it curbs unnecessary exploration which aids in completing the task efficiently.

\subsection{Generalization}
We also explore the generalizability of the proposed PAC-RoManNet architecture by training it using a learning from demonstration framework on the 10 benchmark tasks presented by Zeng \etal in \cite{zeng2020transporter}. The Ravens-10 benchmark tasks are difficult as most methods tend to over-fit to the training demonstration and generalize poorly with less than 100 demonstrations. The performance results in Table \ref{tab: Benchmark} show that PAC-RoManNet can achieve state-of-the-art performance in terms of success rate on Ravens-10 benchmark tasks. While other methods require a hundred or thousand demonstrations to achieve a task success rate of over 90\% for tasks like \textit{packing-boxes} and \textit{sweeping-piles}, PAC-RoManNet requires less than 1/10th of the number of demonstrations. This validates that the sampling efficiency of PAC-RoManNet is extremely impressive when evaluated on unseen test settings.


\section{Conclusions}
In this work, we present a vision-based deep reinforcement learning framework to effectively learn complex manipulation tasks that consist of multiple steps and long-horizon planning. The experimental results indicate that our TPG reward which computes reward based on the actions that lead to successful motion primitives and progress toward the overall task goal can successfully handle progress reversal in multi-step tasks. Moreover, we showed that our LAE policy can be used to curb unnecessary exploration which can occur after the initial states have already been explored. Compared to the previous work in multi-step manipulation, empirical results demonstrate that our method performs significantly better than all previous methods in various multi-step tasks in simulation and the real-world setting. We also performed several experiments for ablation studies to further examine the effects of each component in our system. Finally, we demonstrate a high generalization ability of PAC-RoManNet by showing that it can achieve state-of-the-art performance on Ravens-10 benchmark tasks.


\section*{Acknowledgment}

The authors acknowledge Research Computing \cite{ritrc} at the Rochester Institute of Technology for providing computational resources and support that have contributed to the research results reported in this publication.

\balance
\bibliographystyle{IEEEtran} 
\bibliography{references} 


\end{document}